\documentclass[10pt,twocolumn,letterpaper]{article}

\usepackage[pagenumbers]{cvpr} %

\usepackage{csquotes}
\usepackage{algorithm}
\usepackage{algpseudocode}
\usepackage{adjustbox}

\newcommand{\TODO}[1]{\textbf{\color{red}[TODO: #1]}}
\renewcommand{\TODO}[1]{}

\let\originalcite\cite
\renewcommand{\cite}[1]{}
\newcommand{\scite}[1]{\originalcite{#1}}

\definecolor{cvprblue}{rgb}{0.21,0.49,0.74}
\usepackage[pagebackref,breaklinks,colorlinks,allcolors=cvprblue]{hyperref}

\def\paperID{} %
\def\confName{CVPR}
\def\confYear{2026}
\def\model{DSFlash}

\title{DSFlash: Comprehensive Panoptic Scene Graph Generation in Realtime}

\author{Julian Lorenz \quad Vladyslav Kovganko \quad Elias Kohout\\Mrunmai Phatak \quad Daniel Kienzle \quad Rainer Lienhart\\
University of Augsburg\\
{\tt\small julian.lorenz@uni-a.de}
}

\begin{document}
\maketitle

\begin{abstract}
Scene Graph Generation (SGG) aims to extract a detailed graph structure from an image, a representation that holds significant promise as a robust intermediate step for complex downstream tasks like reasoning for embodied agents.
However, practical deployment in real-world applications - especially on resource constrained edge devices - requires speed and resource efficiency, challenges that have received limited attention in existing research.
To bridge this gap, we introduce DSFlash, a low-latency model for panoptic scene graph generation designed to overcome these limitations. DSFlash can process a video stream at 56 frames per second on a standard RTX 3090 GPU, without compromising performance against existing state-of-the-art methods. Crucially, unlike prior approaches that often restrict themselves to salient relationships, DSFlash computes comprehensive scene graphs, offering richer contextual information while maintaining its superior latency. Furthermore, DSFlash is light on resources, requiring less than 24 hours to train on a single, nine-year-old GTX 1080 GPU.
This accessibility makes DSFlash particularly well-suited for researchers and practitioners operating with limited computational resources, empowering them to adapt and fine-tune SGG models for specialized applications.
\end{abstract}
    
\section{Introduction}
\label{sec:intro}

\begin{figure}[t]
  \centering
  \includegraphics[width=\linewidth]{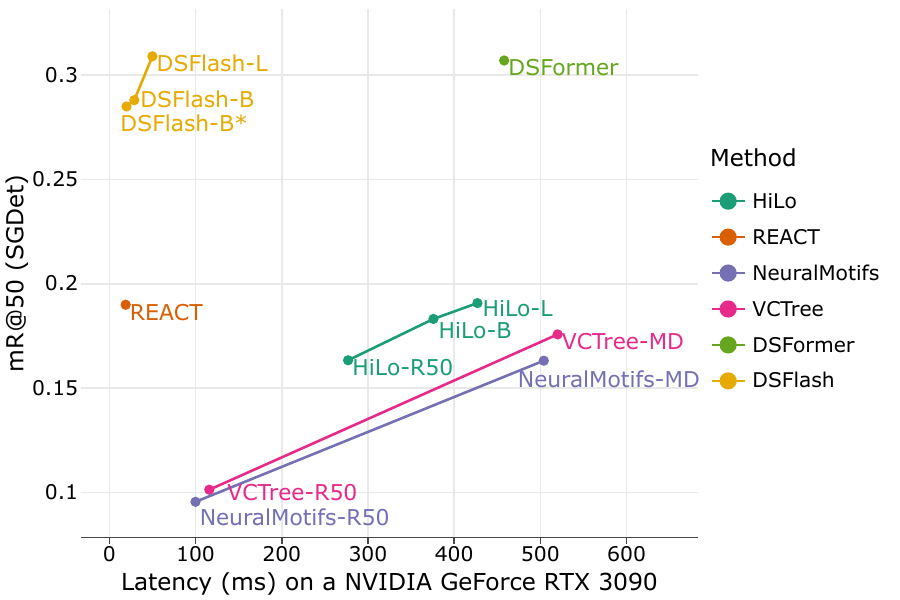}

  \caption{Performance comparison between our approach and previous work in terms of performance (\emph{mR@50}) and latency (ms) on the PSG dataset \cite{psg}.}
  \label{fig:teaser}
\end{figure}

Visual scene understanding has long been a core challenge in computer vision.
One increasingly valuable approach involves scene graph generation (SGG) \cite{first-sg}, which provides a rich, structured representation of an image.
A scene graph is a set of nodes (representing instances in the scene) and edges (representing the relationships between them).
This structure allows models to move beyond simple object detection towards comprehending complex interactions and context within a visual scene.
When talking about relations, we refer to a triplet of (subject, predicate, object) that describes a directed relationship, \eg (\enquote{person}, \enquote{sitting on}, \enquote{chair}).

Scene graphs have proven to be highly beneficial for numerous downstream tasks, including visual question answering \cite{intro-vga1,intro-vqa2,intro-vqa3}, sports analysis \cite{relwork-sgsport1,relwork-sgsport2,relwork-sgsport3}, assisting surgery \cite{relwork_surgery,relwork_surgery2}, image captioning \cite{intro-caption1,intro-caption2,intro-caption3}, and embodied reasoning tasks \cite{intro-reasoning1,intro-reasoning1,intro-reasoning3}.
Crucially, they serve as an explainable intermediate step, offering clear, human-readable insights into a model's decision-making process, contributing a significant advantage over opaque end-to-end models.

SGG methods offer a lightweight and interpretable alternative to complex vision-language models (VLMs) in domains where structured scene understanding is prioritized over broad multimodal reasoning. However, despite their utility, the vast majority of research in this field has focused on improving the quality of extracted scene graphs, often at the expense of computational efficiency. Consequently, almost no research has been dedicated to low-latency scene graph generation models suitable for real-time or resource-constrained applications, with a few recent exceptions \cite{react}.
This gap is critical, as many real-world applications, especially autonomous systems with limited communication, demand efficient, scalable solutions that can operate under strict latency constraints. To the best of our knowledge, there is no active research on low-latency models that are specifically targeted for panoptic scene graph generation (PSGG).

In this paper, we address this gap by presenting a novel, low-latency PSGG model, called \model{}.
Our work aims to demonstrate that highly efficient PSGG is still achievable while maintaining competitive performance.
In fact, we go one step further and predict comprehensive scene graphs, \ie for each image, \model{} localizes all instances and classifies all potential relations between all instances.
Even then, \model{} runs faster than existing SGG methods that only classify a subset of relations.

\textbf{Evaluation scenario.}
To benchmark SGG models for real-world applications, we evaluate model latency with a batch size of 1, simulating a constant feed of frames from a video stream.
For our time measurements, we analyze latency as the time taken for a forward pass through a model.
Since we are also interested in resource-constrained applications, we evaluate \model{} not only using highly optimized hardware but also using older GPUs.

Our contributions can be summarized as:

\begin{enumerate}
  \item We introduce \model{}, a low-latency panoptic scene graph generation method with SOTA performance.
  \item We propose a bidirectional relation predictor that halves the number of forward passes through the model head.
  \item We present a mask-based dynamic patch pruning technique to reduce the number of processed tokens with minimal overhead.
  \item We perform a thorough comparison with other PSGG models regarding performance and latency.
  \item We show an extensive ablation study of the influence of \model{}'s various components on latency and performance.
\end{enumerate}

We will publish our code and extensive collection of model checkpoints upon paper acceptance.

\section{Related Work}

In this section, we discuss related work regarding other SGG methods and relevant components that are employed in \model{}.

\subsection{Panoptic Scene Graph Generation}

Contrary to traditional scene graph generation, panoptic scene graph generation (PSGG) \cite{psg} uses segmentation masks instead of bounding boxes to localize instances in a scene.
Integrating segmentation masks into the training process gives a model additional context and understanding of various relations as opposed to bounding boxes.
The most commonly used dataset for PSGG is the PSG dataset \cite{psg} which was created from the overlap of the Visual Genome \cite{visual-genome} and COCO \cite{coco} datasets, resulting in 49k images with panoptic segmentation masks and scene graph annotations.
Contrary to Visual Genome, PSG reduces the large predicate vocabulary of Visual Genome to 56 carefully selected classes.

The PSGG task has been tackled with various approaches, including methods that were originally built for Visual Genome like NeuralMotifs \cite{motifs} or VCTree \cite{vctree} but also specialized methods like HiLo \cite{hilo} or DSFormer \cite{dsformer}.

Similarly to classical SGG, PSGG methods can be categorized into two-stage and one-stage methods.
Two-stage methods first predict instance locations and categories and use that information to perform the actual relation classification.
One-stage methods do not separate these steps and directly predict a set of subject-predict-object triplets.
Although recent research frequently argues for one-stage methods \cite{integrapsg,hilo,psg} because of their supposedly superior performance and efficiency, there have been recent publications opposing this claim \cite{react,dsformer}.

We choose to follow a two-stage approach for \model{} since it possesses beneficial properties that align with our target setting.
By reusing a pretrained segmentation backbone, \model{} can be trained much quicker and with less resource requirements.
On the other hand, if enough computational resources are available, the segmentation stage can be fine-tuned to specific datasets, even using potential pretraining recipies from adjacent domains.

\subsection{Realtime Scene Graph Generation}

Research on low-latency models for scene graph generation has remained largely overlooked, with only a few studies reporting their latency metrics.

However, \citeauthor{react} \cite{react} have begun to investigate common bottlenecks in recent SGG architectures.
They evaluate multiple existing two-stage architectures \cite{vctree,motifs,TDE} and optimize them for speed, achieving groundbreaking latency with good SGG performance using their new architecture, named REACT \cite{react}.
The most impactful contribution to the low latency is the replacement of the Faster-RCNN \cite{faster-rcnn} backbones with YOLOv8 \cite{yolo8}.
Additionally the authors propose to filter the number of box proposals before classifying relations between them, reducing the number of required relation predictions.

Although certainly a fundamental first step towards real-time efficient SGG models, the PSGG setting is only covered very briefly.
In this work, we present a PSGG model that fully utilizes all available modalities and outperforms REACT \cite{react} by a large margin in latency and PSGG performance.

\subsection{DSFormer}
\label{sec:dsformer}

To design \model{}, we borrow some concepts from DSFormer \cite{dsformer} but replace many parts for increased performance and speed.
The core idea behind DSFormer is to strictly decouple instance segmentation and relation prediction not only into two separate stages of the network but two entirely separate networks.
This enables DSFormer to immediately use any modern segmentation model in a plug-and-play fashion without having to retrain the scene graph model.
Although certainly appealing, having to use two networks with separate backbones involves an immense resource inefficiency which we address with \model{}.

To produce a panoptic scene graph, DSFormer first executes a segmentation model to retrieve segmentation masks.
Next, for each possible combination of two segmentation masks, the second stage is prompted with the image, a subject segmentation mask, and an object segmentation mask.
The model then classifies the relation between subject and object.
To encode the location of subject and object, DSFormer adds a mask embedding to each patch:
\begin{equation}
    token = patch + r_s \cdot t_s + r_o \cdot t_o + (1 - r_s - r_o) \cdot t_{bg},
    \label{eq:sbjobj}
\end{equation}
where $r_s$ and $r_o$ are the proportions of the patch area that is covered by the subject and object mask.
$t_s, t_o, t_{bg}$ are learnable tokens.
Next, the enriched patches are processed by a set of transformer blocks.
Finally, the relation head projects the tokens to the desired relation output.

While DSFormer achieves SOTA results on PSG, efficiency has been completely neglected by the authors.
With \model{} we borrow some core concepts to achieve similar PSGG scores while achieving significantly reduced inference time.

\subsection{Fast Vision Backbones}

As with most neural network architectures that target fast inference, an optimized backbone is key.
A common choice is a network from the YOLO family \cite{yolo1}, with recent iterations \cite{yolo10,yolo11,yolo12} achieving around 40 mAP on bounding box detection in less than 2~ms on a T4 GPU.

Although initially regarded as less efficient than convolutional neural networks, transformer based architectures, especially ViT-based methods have eventually shown superior computational efficiency \cite{survey-fastvit} powered by highly optimized algorithms like flash attention \cite{flashattn,flashattn2}.
Especially during inference, ViTs benefit from data locality and use of optimized matrix operations.
The recently published RF-DETR \cite{rf-detr} model applies these insights. It is based on a transformer architecture, namely LW-DETR \cite{lw-detr} and Deformable DETR \cite{def-detr} and achieves 48 mAP on bounding box detection in 2.3~ms on T4 GPU.

Another resource efficient vision backbone is the Encoder-only Mask Transformer (EoMT) \cite{eomt}.
It achieves detection performance on a par with SOTA models while improving latency.
By using only a Vision Transformer \cite{vit} (ViT) encoder, EoMT eliminates the need for sequential components like feature adapters, pixel decoders, and separate transformer decoders (\eg Mask2Former \cite{mask2former}).
It integrates segmentation directly into the encoder's attention mechanism, enabling parallel processing of image patch and segmentation query tokens in a single forward pass.
This approach achieves up to $4\times$ faster inference than Mask2Former, especially with large models (\eg ViT-L), while maintaining competitive accuracy.
EoMT's efficiency is further enhanced by large-scale self-supervised pretraining (\eg DINO \cite{dinov1,dinov2,dinov3} or EVA-02 \cite{eva02}), which provide the robust visual representations needed for segmentation without auxiliary components.
Its minimalist design makes it ideal for low-latency, real-time applications and easy to integrate into an existing codebase.
We therefore choose EoMT as the backbone for \model{}.

\subsection{Evaluation Issues}
\label{sec:eval-bad}

For a comparison with prior work on PSGG, we use the PSG dataset \cite{psg} to report performance results.
Like most scene graph datasets, PSG contains no negative ground truth annotations that indicate when a certain relation is predicted incorrectly.
Therefore, we have to fall back to ranking metrics like Recall@k (\emph{R@k}) to evaluate model performance.
There are some exceptions that address negative ground truth, however these datasets are either only designed for evaluation \cite{haystack} or based on simulated data only \cite{copasg}.
A well understood problem of using \emph{R@k} for SGG is the long tail distribution of the used dataset \cite{relwork_invarlearn,relwork_multiproto,relwork_semproto,hilo,ietrans,relwork_cktrcm}.
A model that gets only the top 5\% of all predicate classes right would already achieve a \emph{R@k} of 52\%.
For a more balanced metric, we report Mean Recall@k (\emph{mR@k}) instead, which first computes the \emph{R@k} per predicate class using all images and averages the individual scores for the final metric.

When evaluating PSGG (and SGG) methods, special care is required when interpreting the model output from various models.
Recent work \cite{dsformer,react} has investigated irregularities in the evaluation protocol of prior SOTA methods.
Models generated multiple relation predictions for the same associated ground truth subject-object pair.
Consequently, SOTA models artificially increased the number of attempts per ground truth, which is not allowed per \emph{mR@50} definition.

\section{Method}

\begin{figure*}[t]
  \centering
  \includegraphics[width=1.0\linewidth]{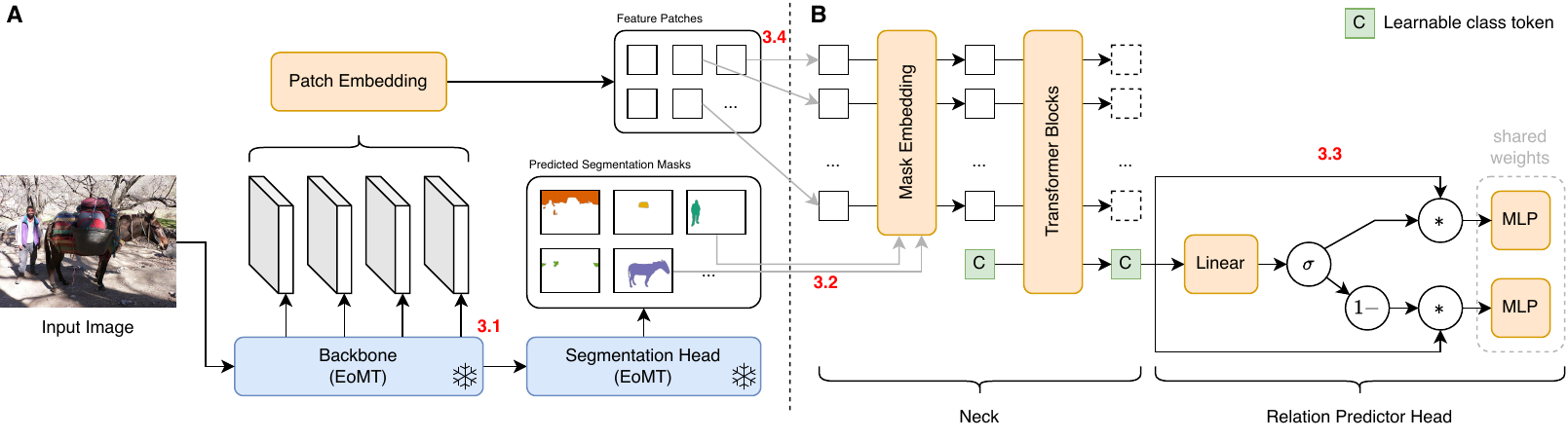}
  \caption{
    Overview of the DSFlash architecture for inference.
    Part A is executed once per image.
    Part B is executed for each combination of two segmentation masks.
    We use EoMT \cite{eomt} as the segmentation backbone which is kept frozen throughout the whole training.
    We use the mask embedding module from DSFormer \cite{dsformer}.
    The relation predictor head is described in \cref{sec:bidir}.
    The red numbers indicate which components are covered in which section.
  }
  \label{fig:arch-backbone}
\end{figure*}

We use DSFormer \cite{dsformer} as the baseline for our efficient architecture and provide an overview over the relevant parts in the supplementary.
We choose this model as our baseline because of its SOTA performance and simple design, allowing us to quickly iterate new optimizations into the code.
Starting from DSFormer, we re-create several parts of the model and end up with a new architecture that is capable to run PSGG at very high speeds while producing more accurate relation predictions.

An overview of the final \model{} architecture is shown in \cref{fig:arch-backbone}.
To process a single image, \model{} first uses a pretrained EoMT \cite{eomt} backbone to extract feature patches and predict a comprehensive set of segmentation masks.
For every combination of two segmentation masks (\eg a person and a donkey), a mask embedding \cite{dsformer} is added to the feature patches from the backbone, encoding the location of subject and object.
The enriched patches are then fed through a set of tranformer blocks (model neck) and finally processed by a relation head, predicting the associated relations for the two masks in both directions simultaneously, \eg \enquote{person behind donkey} and \enquote{donkey right of person}.

Note that during training, DSFlash receives ground truth segmentation masks instead of inferred segmentation masks.
This provides better supervision through better segmentation masks and reduces training time by skipping the segmentation head.

\subsection{Merged Backbones}
\label{sec:backbone}

Starting off the architecture of DSFormer, we initially inherit its strictly decoupled approach, meaning that segmentation is delegated to a separate upstream model.
Since DSFormer requires the inferred segmentation masks as input, an additional forward pass through a segmentation model is required per frame.
Still, DSFormer contains its own ResNet-based backbone which results in costly forward passes through two backbones.

To tackle this inefficiency, we decide to directly extract the feature tensors from the segmentation model, which already runs to produce the segmentation masks.
In addition, we replace the slow MaskDINO \cite{maskdino} segmentation model with EoMT \cite{eomt} which produces panoptic segmentation masks of similar quality at a much lower latency.
To get a spatial feature tensor from the backbone, we extract tokens after blocks 2, 5, 8, and 11 for the S and B variants and blocks 5, 11, 17, 23 for the L variant.
Since EoMT uses DINO \cite{dinov2,dinov3}, we make sure to only extract patch tokens and exclude the class and register tokens.
After block 9 (large: 20), EoMT introduces additional query tokens which we also exclude.
Concatenating the extracted tokens along the embedding dimension gives us a feature tensor with the shape $768 \times 40 \times 40$ for a $640 \times 640$ input image.
\cref{fig:arch-backbone} illustrates the approach.

We keep the EoMT-based backbone \emph{frozen at all times}.
This speeds up training and enables us to fully outsource panoptic segmentation training which can be performed much more efficiently and on larger datasets without the scene graph generation context.

\subsection{Raw-resolution Segmentation Masks}
\label{sec:lowres}

To optimize computational efficiency in the later stages of the architecture, we adopt a ViT-style patch embedding \cite{vit}, splitting the $40 \times 40$ feature tensor into $13 \times 13$ patch tokens, each with an embedding dimension of 384.
Building on DSFormer, DSFlash integrates the extracted segmentation masks directly into these patch tokens. Specifically, a weighted embedding is added to each patch, where the weight is determined by the fraction of the patch's area that is covered by the corresponding segmentation mask (as formalized in \cref{eq:sbjobj}).
A more in-depth description can be found in the supplementary and the original paper \cite{dsformer}.

\begin{figure}[t]
  \centering
  \begin{subfigure}{0.495\linewidth}
    \includegraphics[width=\linewidth]{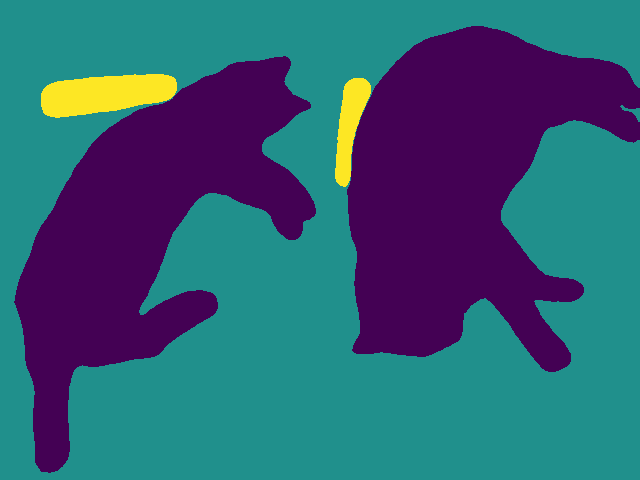}
    \caption{Upsampled mask logits}
  \end{subfigure}
  \hfill
  \begin{subfigure}{0.495\linewidth}
    \includegraphics[width=\linewidth]{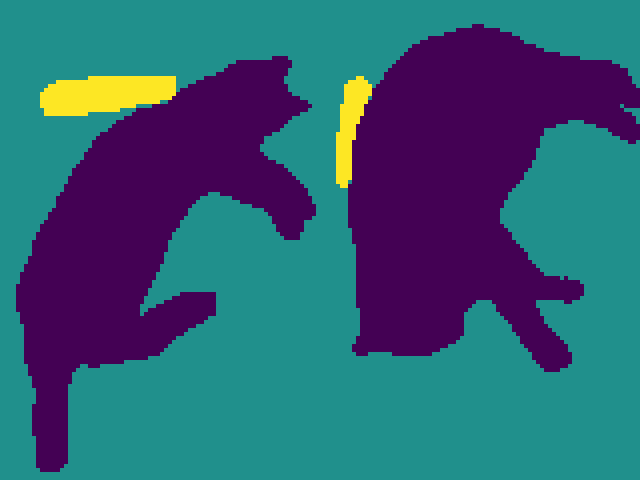}
    \caption{No upsampling}
  \end{subfigure}
  \caption{Qualitative comparison of the segmentation masks produced by EoMT with and without upsampling the logits.}
  \label{fig:hilo-res}
\end{figure}

Following DSFormer, we would use the segmentation mask logits from EoMT ($160 \times 160$), upsample them to the image size and then compute the patch area overlap fractions.
However, since only a resolution of $13 \times 13$ is required to compute the embedding, there is no need to upsample to the image size, and we can skip the costly bilinear interpolation step.
\cref{fig:hilo-res} visualizes the difference between the different resolutions.

\subsection{Bidirectional Predictions}
\label{sec:bidir}

\begin{figure}[t]
  \centering
  \includegraphics[width=1.02\linewidth]{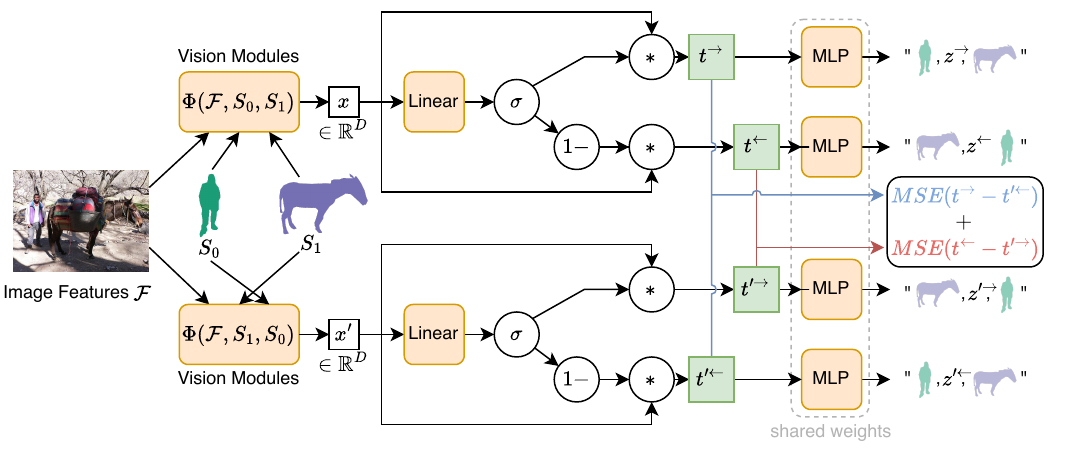}
  \caption{
    Schematic of DSFlash's gating mechanism and the enforced feature consistency loss during training.
    Given two segmentation masks and an image, DSFlash computes a class token $x$ using various modules, summarized here as $\Phi$.
    To train the consistency loss, DSFlash performs two forward passes through the model head with flipped segmentation masks.
  }
  \label{fig:bidir}
\end{figure}

To detect potential relations between two instances in the scene represented by their respective segmentation masks $S_0$ and $S_1$, the original DSFormer has to perform two forward passes through the model neck: one to classify the relation with $S_0$ as the subject and $S_1$ as the object and another one for the opposite.
To reduce the computational cost during inference, we design \model{} such that it encodes both directions in one single forward pass.
This allows us to half the number of predictions required to build a comprehensive scene graph, while keeping the number of parameters in the model almost the same, as we will show in the following.
A schematic can be found in \cref{fig:arch-backbone} (relation predictor head) and \cref{fig:bidir}.

Let's assume that the backbone has extracted two segmentation masks $S_0$ and $S_1$ for a given image with feature tensor $\mathcal{F}$.
The task is now to classify two potential relations.
We call the one where $S_0$ is the subject and $S_1$ is the object the \emph{forward} prediction $z^\rightarrow \in \mathbb{R}^C$ and the reversed one the \emph{backward} prediction $z^\leftarrow \in \mathbb{R}^C$, with $C$ being the number of predicate classes.
Accordingly, $y^\rightarrow, y^\leftarrow \in \{0, 1\}^C$ are the associated target ground truth values.

Both DSFormer and \model{} first encode the masks into the image feature tensor $\mathcal{F}$ using a specialized mask embedding module \cite{dsformer}, resulting in an enriched feature tensor $x = embed(\mathcal{F}, S_0,  S_1) \in \mathbb{R}^D$, with embedding dimension $D$.
DSFormer, then processes $x$ in its prediction head to produce $z^\rightarrow$.
For $z^\leftarrow$, DSFormer computes a second forward pass with $x' = embed(\mathcal{F}, S_1,  S_0)$ since $embed$ is not symmetrical.

\model{} on the other hand uses $x$ for both directions and produces $z^\rightarrow$ and $z^\leftarrow$ in a single forward pass.
Before processing $x$ with a relation predictor head, we split $x$ into two intermediate tensors $t^\rightarrow$ and $t^\leftarrow$ using a gate mechanism, described in \cref{eq:g1,eq:g2,eq:g3} which is loosely inspired by Gated Recurrent Units \cite{gru}.
Finally a shared MLP predicts the relations (\cref{eq:g4,eq:g5}).
\begin{align}
  \label{eq:g1}
  g &= \sigma\left(gate_{mlp}(x)\right) \in \mathbb{R}^D \\
  \label{eq:g2}
  t^\rightarrow &= g \odot x \in \mathbb{R}^D \\
  \label{eq:g3}
  t^\leftarrow &= (1 - g) \odot x \in \mathbb{R}^D \\
  \label{eq:g4}
  z^\rightarrow &= relhead_{mlp} (t^\rightarrow) \in \mathbb{R}^C \\
  \label{eq:g5}
  z^\leftarrow &= relhead_{mlp} (t^\leftarrow) \in \mathbb{R}^C
\end{align}

To learn the relation predictions, we follow DSFormer and use a binary cross entropy loss function between $(z^\rightarrow, y^\rightarrow)$ and $(z^\leftarrow, y^\leftarrow)$.

In our initial experiments we sorted $S_0$ and $S_1$ by their ordering in the ground truth.
The model quickly discovered that a positive annotation is 3 times more likely to occur in $y^\rightarrow$ than in $y^\leftarrow$ in the PSG dataset, making the model rely on a fact that cannot be determined in a real-world application.
To prevent this, we introduce the shared relation predictor from \cref{eq:g4} and add an additional feature consistency loss, shown in \cref{eq:consistency}.

During training, we perform two forward passes through DSFlash's model head for each $(S_0, S_1)$ pair with swapped ordering of the segmentation mask when embedding them into $\mathcal{F}$.
This gives us a second enriched feature tensor $x' = embed(\mathcal{F}, S_1, S_0) \neq x$.
From here, we compute $t'^\rightarrow, t'^\leftarrow, z'^\rightarrow, z'^\leftarrow$ analogously to \cref{eq:g2,eq:g3,eq:g4,eq:g5}.
Ideally, DSFlash would predict $z^\rightarrow = z'^\leftarrow$ and $z^\leftarrow = z'^\rightarrow$ because $S_0$ and $S_1$ are flipped.
To assist the model to learn this similarity, we use an additional consistency loss on the intermediate tensors:
\begin{align}
  \text{Consistency} = \frac{1}{D} \sum_{i=1}^{D} (t^\rightarrow_i - t'^\leftarrow_i)^2 + (t^\leftarrow_i - t'^\rightarrow_i)^2
  \label{eq:consistency}
\end{align}

During inference, DSFlash can rely on its learned ability to treat each directions equally and only a single forward pass is required.

\subsection{Mask-Based Dynamic Patch Pruning}
\label{sec:dyn-pruning}

The time spent in the model neck is influenced by the number of processed patch tokens.
To reduce the computational cost, we introduce a dynamic patch pruning strategy that is tailored towards \model{}.

Before the extracted patch tokens from the backbone enter the model neck, an additional subject-object mask embedding is added to each patch token.
This mask embedding is directly determined by the overlap ratio of a mask with the respective patch area.
Consequently, a patch that is neither overlapping with the subject nor with the object, receives no additional embedding.
Although they provide some context of the overall scene, these patches contain almost no helpful information to determine the final relation predicates.
For an additional speedup, we therefore identify the patches that have no overlap with subject or object and drop them before processing them with the model neck.
Since the overlaps are computed anyway, the patches to prune can be identified with virtually no computational overhead.

Since the final predictions only rely on information of the classification token, \model{} can handle variable token numbers.

\subsection{Token Merging}
\label{sec:tome}

We apply token merging \cite{tome} to reduce the computational cost when calculating attention in the transformer layers.
Specifically, we use ToMe-SD \cite{seg++,tomesd} because it unmerges the tokens again which better retains the segmentation capabilities of the backbone.
Before every attention layer in \model{}'s backbone, we use ToMe-SD to merge similar tokens.
Directly after the attention layer, ToMe-SD unmerges the tokens to the initial number.

\subsection{Additional Improvements}
\label{sec:m:additional}

To further reduce latency and increase scene graph performance of \model{}, we perform several improvements.

\textbf{Efficient subject/object mask encoder.}
We observe that DSFormer's mask encoder contains inefficient code that involves several copies of the extracted segmentation masks followed by multiple tensor operations that can be exactly represented by an efficient average pooling layer.
This is described in more detail in the supplementary.

\textbf{DeiT III-style augmentation.}
We choose to follow the data augmentation strategy from DeiT III \cite{deit3}. We apply random horizontal flips and color jitter, then randomly select one of three transformations: grayscale conversion, solarization, or Gaussian blur.

\section{Results}
\label{sec:results}

We start with introducing the evaluation metrics and continue with a comparison with other PSGG methods on the PSG dataset.
We then analyze various components of DSFlash for the impact on the final performance and speed.

\subsection{Metrics}

\textbf{Latency.}
In \cref{sec:results}, we refer to latency as the average time it takes a PSGG model to process a single image.
This evaluates a model's capability to be applied in a real-world environment where image frames have to be processed sequentially from a continuous video stream.
To ensure a fair comparison, we only measure the time of a model's forward pass without any data preprocessing steps.
All latency values are measured using a batch size of 1 to simulate the single image scenario.
We wait for a warmup phase of 200 forward passes before starting the time measurements and report the average latency in the tables and figures.

\textbf{RPS.}
Since neural networks are highly optimized for modern GPUs, various speed improvements are only noticable if the GPU is working at maximum capcity.
Therefore, we also evaluate a model's speed when processing larger batches of data instead of just single-image latency.
We report Relations per Frame (\emph{RPS}) as the average number of subject-object pairs that can be processed by the evaluated model per second.

\textbf{mR@50.}
To measure scene graph performance, we report \emph{mR@50} \cite{vctree,kern} which handles the predicate class imbalances in scene graph datasets much better than other metrics like \emph{R@k}.
As highlighted in \cref{sec:eval-bad}, we make sure that multiple mask predictions per ground truth and multiple relation predictions per ground truth are prevented.
We evaluate all models using the Scene Graph Detection (SGDet) protocol, sometimes also refered to as SGGen.
With this protocol, a PSGG model only receives an image as input and must predict correct segmentation masks, instance labels and subject-predict-object triplets.
Additional results with the PredCls protocol are in the supplementary.

\subsection{Main Results}

\begin{table}
  \caption{
    Performance comparison on the PSG dataset \cite{psg}.
    All models are evaluated using a batch size of 1 on an RTX 3090 GPU and the SGDet protocol.
    Models marked with a star (*) use low-resolution segmentation masks, discussed in \cref{sec:lowres}
  }
  \label{tab:main-results}
  \centering
  \begin{tabular}{@{}lrrr@{}}
    \toprule
    Method & mR@50 $\uparrow$ & Latency $\downarrow$ & \# Params \\
    \midrule
    MotifNet-R50 & 9.56 & 100 ms & 109M \\
    MotifNet-MD & 16.32 & 504 ms & 332M \\
    VCTree-R50 & 10.14 & 116 ms & 105M \\
    VCTree-MD & 17.58 & 520 ms & 327M \\
    HiLo-R50 & 16.34 & 277 ms & 59M \\
    HiLo-L & 19.08 & 427 ms & 230M \\
    REACT & 19.00 & 19 ms & 43M \\
    DSFormer & 30.70 & 458 ms & 330M \\
    \midrule
    DSFlash-L & \textbf{30.90} & 50 ms & 340M \\
    DSFlash-B* & 28.50 & 23 ms & 116M \\
    DSFlash-S* & 25.05 & \textbf{18 ms} & 40M \\
    \bottomrule
  \end{tabular}
\end{table}

A visual comparison of various PSGG methods can be seen in \cref{fig:teaser}, with more detailed results presented in \cref{tab:main-results}.
\model{} outperforms all other methods on \emph{mR@50}, even DSFormer by a small margin of 30.9 vs 30.7.
At the same time, the largest \model{} variant which uses an EoMT-L backbone is still faster than all other models with the exception of REACT.
Although \model{}-L has the most parameters in our evaluation, its simple architecture allows it to run efficiently on optimized hardware.
On the other side, \model{}-S* uses an EoMT-S backbone combined with low-resolution segmentation masks, discussed in \cref{sec:backbone}, enabling it to run with a latency of 18~ms on a NVIDIA GeForce RTX 3090 GPU, equivalent to a frame rate of 56 frames per second.
With only 40M parameters, it is also the smalles model in our comparison, while still achieving superior \emph{mR@50} scores compared to all other methods, except DSFormer.

We attribute \model{}'s efficiency to the use of a modern backbone with good hardware support, as well as our simple segmentation head that only uses optimized tensor operations.

\subsection{Backbone Ablation}
\label{sec:bb-abl}

\begin{figure}[t]
  \centering
  \includegraphics[width=0.9\linewidth]{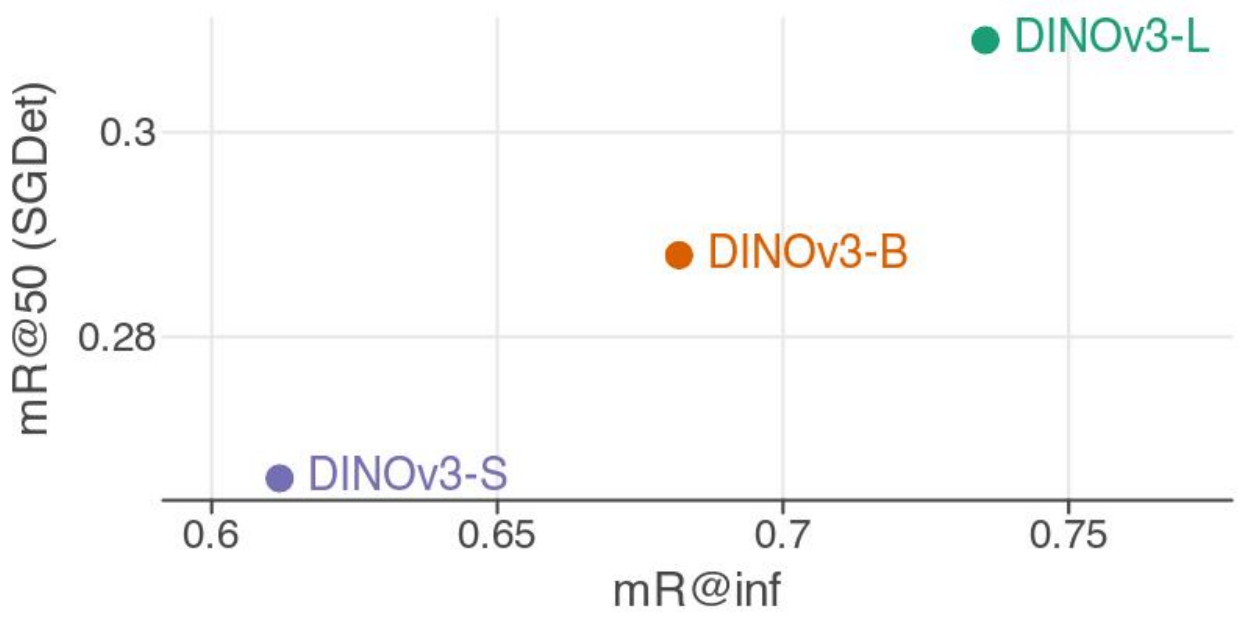}
  \caption{
    Impact of the segmentation model's capability on the final performance.
    \emph{mR@inf} \cite{sgbench} is the best possible \emph{mR@k} that a hypothetical perfect PSGG model could achieve, given the extracted segmentation masks from the segmentation model.
  }
  \label{fig:backbone-impact}
\end{figure}

As shown in previous literature on two-stage methods, the first segmentation stage has a high impact on the final scene graph performance \cite{dsformer}.
We evaluate three training runs with equal settings except for the size of the backbone.
As shown in \cref{fig:backbone-impact}, a larger backbone, is directly correlated to the overall performance.
Given the extracted segmentation masks and associated predicted class labels, we report \emph{mR@inf} \cite{sgbench} as the upper limit to \emph{mR@k} for a hypothetical perfect PSGG which has only access to the segmentation masks from the first stage.
\cref{fig:backbone-impact} shows a striking correlation between \emph{mR@50} and \emph{mR@inf}.
In the supplementary, we also show that \emph{mR@50} is directly correlated to the panoptic quality of the segmentation model.
As more improved segmentation models are developed in the future, it is very likely that \model{} can directly leverage their capability for improved scene graph performance.

\subsection{\model{} Improvements}
\label{sec:r:parts}

An overview of the most impactful optimizations for \model{} can be found in \cref{tab:parts-latency}.
Since we copied many parts from DSFormer, we use it as the baseline model.
As expected, the most effective optimization is the use of a single efficient segmentation model instead of two separate slow models, reducing the latency to 41~ms -- a 10th of the initial time.
At the same time, \emph{mR@50} drops considerably, mostly because DSFlash now works with faster but lower quality segmentation masks form EoMT-3B instead of MaskDINO.

DSFlash's efficient mask embedding module (\cref{sec:m:additional}) reduces latency to 37~ms without affecting \emph{mR@50}.
Our new implementation involves much less data copies, mainly resulting in decreased VRAM usage.

Another speed boost is achieved by performing bidirectional predictions (\cref{sec:bidir}), halving the amount of required forward passes through the model neck, which is reflected in the number of processed relations per second on batched data (\emph{RPS}).
Since most images from the PSG dataset contain only few objects, the number of forward passes is rarely a bottleneck when measuring single-image latency on a GPU.
In addition, our bidirectional training gives the model additional supervision about both relation directions, resulting in an improved \emph{mR@50}.

\begin{table}
  \caption{
    Impact of the optimizations on the overall latency, measured on a NVIDIA GeForce RTX 3090 GPU and a batch size of 1.
    We also report \emph{RPS} as the processed relations per second when processing batched data.
    The rows are read from top to bottom with each row adding an incremental optimization to the evaluated model.
    The last row is an exception and is an improvement from the row marked with $^1$.
  }
  \label{tab:parts-latency}
  \centering
  \begin{tabular}{@{}lrrr@{}}
    \toprule
    Method & mR@50 $\uparrow$ & Latency (ms) $\downarrow$ & RPS $\uparrow$ \\
    \midrule
    Baseline & 30.7 & 445 & 435 \\
    Unified Backbone & 25.0 & 41 (-91\%) & 5,745 \\
    Eff. Mask Embed & 25.0 & 37 (-10\%) & 7,132 \\ %
    Gated Bidir. Pred.$^1$ & 28.8 & 29 (-22\%) & 11,491 \\ %
    No Seg. Upscaling & 28.5 & 23 (-21\%) & 12,928 \\
    EoMT-3S & 25.1 & 18 (-22\%) & 17,897 \\
    \midrule
    $^1$EoMT-3L & 30.9 & 50 (+72\%) & 5,996 \\ %
    \bottomrule
  \end{tabular}
\end{table}

\begin{figure}[t]
  \centering
  \includegraphics[width=0.9\linewidth]{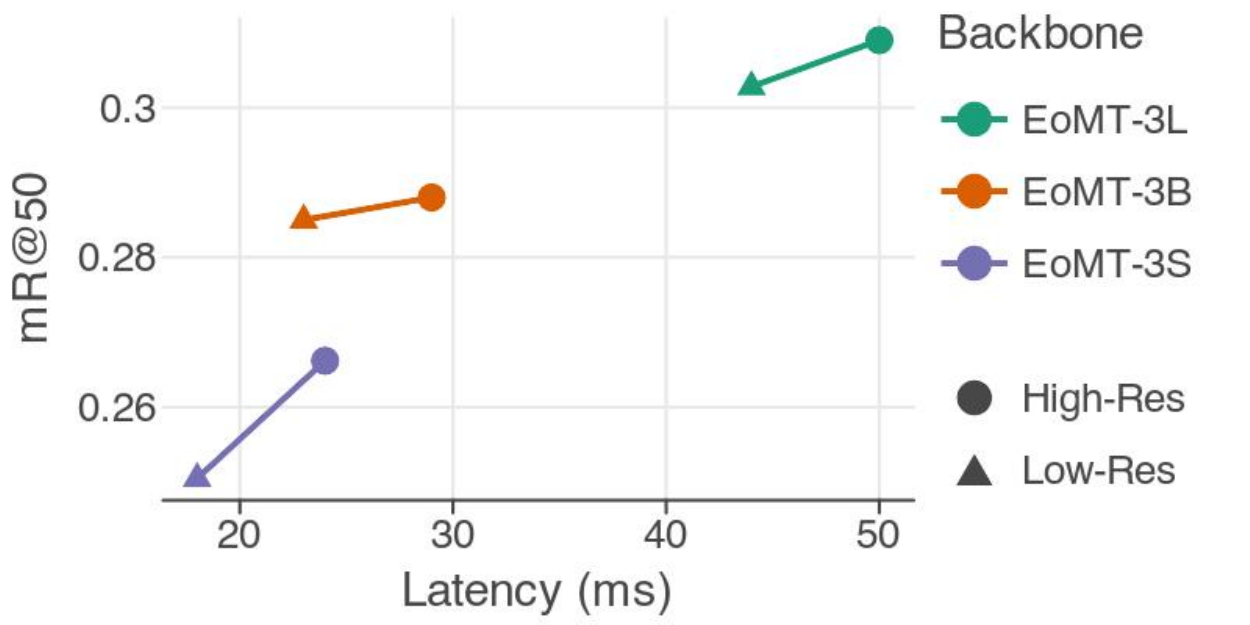}
  \caption{
    Comparison of the effect of low-resolution segmentation masks on the final performance and latency when using different segmentation backbones.
  }
  \label{fig:backbone-lowres}
\end{figure}

A very effective technique to reduce latency is to use low-resolution segmentation masks from the segmentation model.
This skips a costly bilinear interpolation while providing the same granularity of information to the model neck, as discussed in \cref{sec:backbone}.
However, with lower resolution segmentation masks, the final scene graph will also have lower quality and therefore fail to localize some subject and objects from the ground truth.
The impact of low-resolution segmentation masks is also dependent on the capability of the backbone, as illustrated in \cref{fig:backbone-lowres}.
Compared to \model{} models with EoMT-B and EoMT-L backbones, the smaller EoMT-S suffers most from lowered segmentation quality.
An interesting observation is the fact that a model using EoMT-B with low-resolution segmentation masks runs faster and performs better than a model using EoMT-S with high-resolution segmentation masks.
We attribute this to EoMT-S's considerably weaker segmentation capabilities, which directly propagate through the model.
It is worth noting that in some scenarios, the EoMT-S-based model might still be preferred since \model{} with EoMT-S requires only a third of the parameters that \model{} with EoMT-B needs.

\begin{table}
  \caption{
    Impact of Token Merging (\emph{ToMe}) and mask-based dynamic patch pruning (\emph{Prune}, \cref{sec:dyn-pruning}) on \model{}'s latency with batch size 1, measured on a H100, RTX 3090, and GTX 1080 GPU.
  }
  \label{tab:tome}
  \centering
  \begin{tabular}{@{}crrrrr@{}}
\toprule
Prune & ToMe & H100 & 3090 & 1080 & mR@50 \\
\midrule
\textcolor{red}{$\times$} & 0\% & 19 ms & 29 ms & 230 ms & 28.80 \\
\textcolor{Green}{\checkmark} & 0\% & 20 ms & 29 ms & 205 ms & 26.67 \\
\textcolor{Green}{\checkmark} & 30\% & 20 ms & 30 ms & 173 ms & 26.51 \\
\textcolor{red}{$\times$} & 40\% & 21 ms & 29 ms & 180 ms & 25.82 \\
\textcolor{red}{$\times$} & 50\% & 20 ms & 29 ms & 167 ms & 24.87 \\
\textcolor{red}{$\times$} & 60\% & 21 ms & 29 ms & 155 ms & 21.93 \\
\bottomrule
\end{tabular}

\end{table}

\subsection{Pruning and Merging Methods}
\label{sec:drop}

We evaluate the impact of dynamic patch pruning (\cref{sec:dyn-pruning}) and token merging (\cref{sec:tome}) on latency and scene graph performance in \cref{tab:tome}.
These techniques complement each other well since they optimize different parts of the \model{} model.
While dynamic patch pruning reduces the number of patch tokens that enter the model neck, we apply token merging to the attention layers of the segmentation backbone.

We evaluate latency using three GPUs: H100, RTX 3090, and GTX 1080.
We observe that using the more capable H100 and RTX 3090 GPUs results in no notable latency improvement with a batch size of 1.
But when evaluating on batched input, \emph{RPS} still increases.
The exact numbers can be found in the supplementary.
This is most likely because the GPUs are highly optimized for parallel processing and can process all available tokens simultaneously even before pruning.
Lower-end GPUs can process less tokens simultaneously.
Therefore, reducing the parallel workload via pruning has even a measurable impact on reduced latency with batch size 1.

When using dynamic patch pruning (\cref{sec:dyn-pruning}), we notice only a slight performance decrease in \emph{mR@50} of 2.13.
Even though we have not trained it with pruned patches, \model{} still manages to handle the altered input to the model neck.
For additional results where we have trained \model{} with dynamic patch pruning, please refer to the supplementary.
The most notable latency improvement can be observed for a 1080 GPU, where latency drops from 230~ms to 205~ms.

A very promising optimization that reduces latency without deteriorating scene graph performance too much is the combination of token merging and dynamic patch pruning.
Since both optimization methods affect different parts of the architecture, their latency improvements add up.
On a GTX 1080, latency drops from 230~ms to 173~ms while \emph{mR@50} is very similar to our experiment with only patch pruning applied.
Additionally, using a token merging ratio of 30\% combined with patch pruning, gives a similar latency to 50\% token merging without patch pruning, while achieving much better \emph{mR@50}.

\section{Conclusion}

In this work, we have introduced \model{}, a panoptic scene graph generation model that is capable to run with a latency of down to 18 ms on an RTX 3090 GPU while still producing high quality scene graphs.

We have introduced several architectural optimizations and analyze their effect on scene graph quality and latency.
While some techniques like mask-based dynamic patch pruning make direct use of \model{}'s architectural design to reduce latency with minimal overhead, we also introduce several methods that can be easily applied to future methods.
For example, bidirectional relations enable a scene graph model to predict a scene graph while halving the amount of required forward passes, which directly relates to model throughput on batched inputs.
We also experiment with token merging which can be applied to virtually any transformer-based model.
Our experiments show that token merging can be effectively used on resource constrained GPUs.

With the increasing popularity of VLMs in recent research, many systems have grown complex and heavy, often denying the ability to run them on-premise which involves privacy concerns in some areas.
While many applications require the complex reasoning and fundamental scene understanding capabilities, other areas might also be approachable using scene graphs as an efficient intermediate step.
Combined with \model{}'s near instantaneous high quality scene graphs, we hope that future research further explores ressource efficient systems that can be used as an alternative to the influx of overly complex models.

\clearpage
\setcounter{page}{1}
\maketitlesupplementary

\section{Caveats When Evaluating PSGG Methods}
\label{sec:supp:eval-recap}

Recent work \scite{dsformer} has discussed a critical flaw in the evaluation protocol of recent PSGG methods.
In our paper, we follow their introduced SingleMPO protocol which ensures a correct and fair comparison across models.

\begin{figure}[h]
  \centering
  \includegraphics[width=\linewidth]{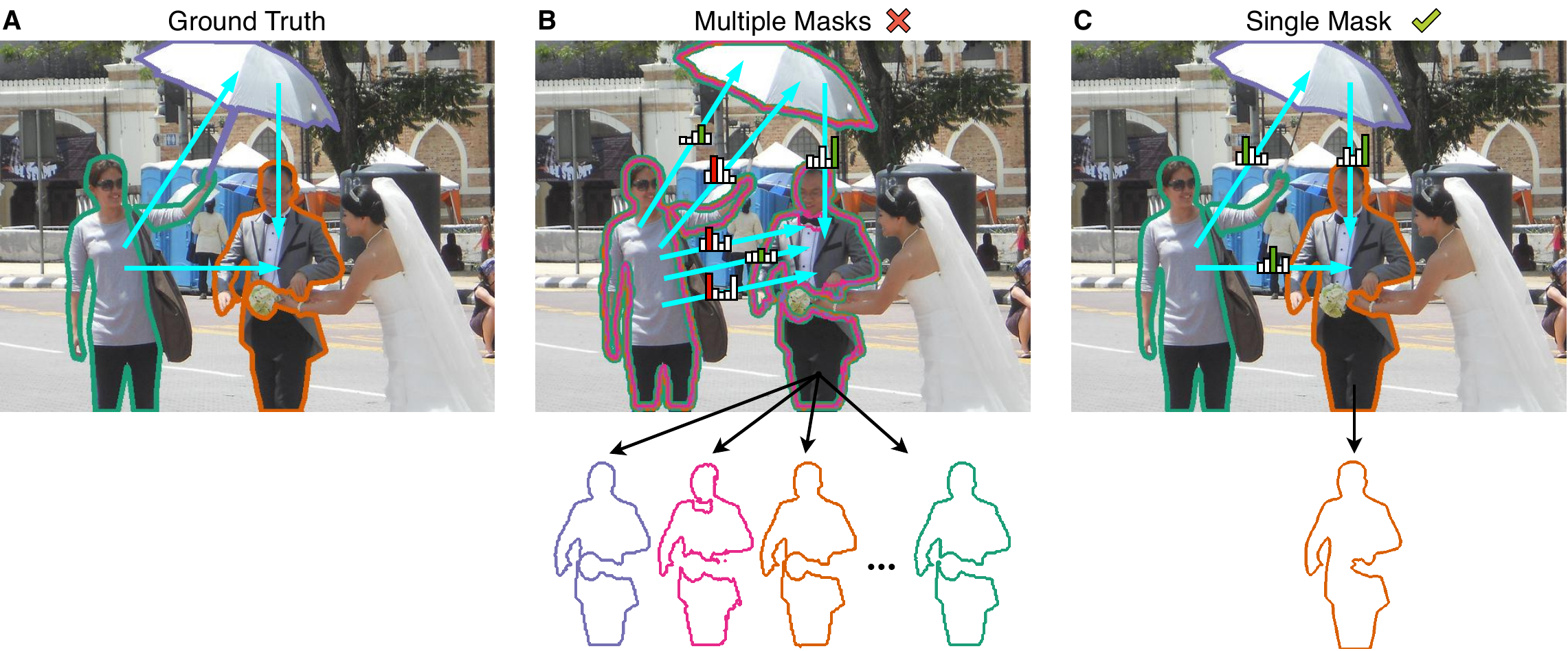}
  \caption{
    Illustration of the impact of multiple masks for the same ground truth mask.
    The model in figure \textbf{B} predicts multiple very similar masks together with separate relation predictions.
    However, this gives the model multiple attempts to predict the ground truth relation, essentially ignoring the definition for \emph{mR@k}.
    Adapted from \scite{dsformer}.
  }
  \label{fig:supp:eval-issue}
\end{figure}

Some recent methods that report SOTA performance on PSGG, output multiple masks for the same object, as illustrated in \cref{fig:supp:eval-issue}.
During evaluation, these methods treat the multiple masks as individual instances and therefore don't merge them.
Consequently, multiple relation triplets are predicted for the same ground truth subject-object pair.
Per definition of the \emph{mR@k} metric, there is only one prediction per subject-object pair allowed.
The faulty implementations do recognize this and ensure that there is only one relation prediction per predicted subject-object pair.
However, since the models essentially predict the same subject-object pair multiple times with slightly different masks, their implementation treats them as separate subject-object pairs which are not filtered.
Consequently, the model gets multiple attempts to correctly predict the ground truth relation.

For DSFlash, we ensure that multiple masks per subject/object are prevented.
This ensures that we don't circumvent the definition of the \emph{mR@k} metric.

\section{DSFormer Mask Embedding}
\label{sec:supp:embed}

Two-stage SGG/PSGG methods require information about the location of subject and object boxes/masks.
A very common approach is to use RoI align \scite{faster-rcnn} to extract spatial features from the associated region in the feature tensor.

DSFormer replaces this approach with an approach somewhat similar to positional embedding.
For each subject-object pair, the model computes an embedding mask that is added to the spatial tokens.
This embedding mask is derived from the overlap ratios of the patches.

\begin{figure}[h]
  \centering
  \includegraphics[width=\linewidth]{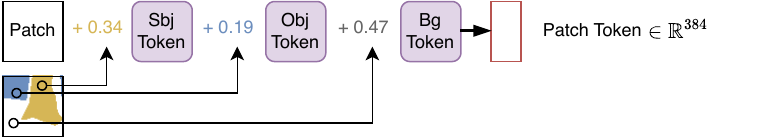}
  \caption{
    Illustration of the mask embedding.
    Subject, object, and background tokens are learnable tokens that are added to the patch embedding with a weighted sum.
    The weights are determined from the proportion of the patch area that is covered by the respective segmentation mask.
    Adapted from \scite{dsformer}.
  }
  \label{fig:supp:patchemb}
\end{figure}

\section{Efficient Mask Embedding}
\label{sec:supp:efficient-maskemb}

\begin{figure}[h]
  \centering
  \includegraphics[width=\linewidth]{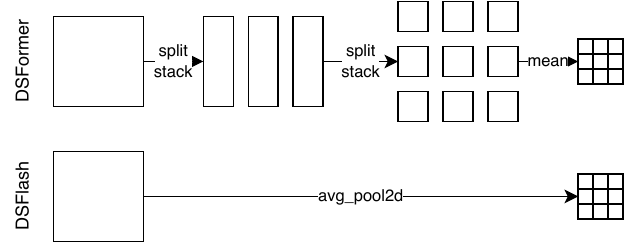}
  \caption{DSFormer's inefficient ratio computation can be represented by a simple average pooling layer.}
  \label{fig:supp:embed-ratios}
\end{figure}

DSFormer's mask embedder implementation uses a convoluted sequence of PyTorch operations including stack, split, flatten, and mean to compute the patch overlap ratios.
As \cref{fig:supp:embed-ratios} illustrates, this sequence can be simplified by using a simple average pooling layer.

As discussed in \cref{sec:supp:embed}, we need to compute the overlap ratios for every subject-object pair.
To this end, DSFormer first copies the segmentation mask for each pair and then compute the ratios for each pair, as illustrated in \cref{alg:supp:theirs}.
For DSFlash, we first compute the ratios for all segmentation masks and then copy the ratios for each pair instead, depicted in \cref{alg:supp:ours}.

\begin{algorithm}
\caption{DSFormer's mask embedding implementation}\label{alg:supp:theirs}
\begin{algorithmic}
\Require image shape ($H, W$), target shape ($H', W'$), number of masks $N$, binary masks $S \in \{0,1\}^{N \times H \times W}$, pairs $P \subset \{1, \dots N\}^2$
\\
\State $E \gets \emptyset \subset \mathbb{R}^{H' \times W'}$
\For{$p \in P$} \Comment Performed in parallel
  \State $s_0 \gets S_{p_0}$ \Comment Allocates additional memory
  \State $s_1 \gets S_{p_1}$
  \State $r_0 \gets pool(s_0) \in \mathbb{R}^{H' \times W'}$
  \State $r_1 \gets pool(s_1) \in \mathbb{R}^{H' \times W'}$
  \State $E \gets E \cup \{make\_embedding(r_0, r_1)$\}
\EndFor
\State \textbf{return} $E$
\end{algorithmic}
\end{algorithm}

\begin{algorithm}
\caption{Our mask embedding implementation}\label{alg:supp:ours}
\begin{algorithmic}
\Require image shape ($H, W$), target shape ($H', W'$), number of masks $N$, binary masks $S \in \{0,1\}^{N \times H \times W}$, pairs $P \subset \{1, \dots N\}^2$
\\
\State $E \gets \emptyset \subset \mathbb{R}^{H' \times W'}$
\State $R \gets pool(S) \in \mathbb{R}^{H' \times W'}$
\For{$p \in P$} \Comment Performed in parallel
  \State $r_0 \gets R_{p_0} \in \mathbb{R}^{H' \times W'}$
  \State $r_1 \gets R_{p_1} \in \mathbb{R}^{H' \times W'}$
  \State $E \gets E \cup \{make\_embedding(r_0, r_1)$\}
\EndFor
\State \textbf{return} $E$
\end{algorithmic}
\end{algorithm}

Since ratios are reused across different pairs, this saves us many calls to the pooling layer while also having to copy less data since the ratio tensors are smaller ($W' < W$ and $H' < H$) than the segmentation mask tensors.
While DSFormer copies segmentation masks and computes ratios $2|P|$ times, DSFlash computes $N$ ratios and only copies the ratios $2|P|$ times.

\TODO{anything else that we should recap?}

\section{Implementation Details}

We use the AdamW optimizer with a learning rate of $10^{-5}$, weight decay of 0.02, and a cosine annealing scheduler with linear warmup.
During training, we clip the gradient norm to a maximum of 1.
To achieve best results, we train for 20 epochs.

For five ground truth relation, we sample one negative relation from the dataset, \ie mask-mask pairs that have no annotation in the ground truth.

In addition to the patch tokens and the classification token, DSFlash also uses the same location token from DSFormer.
We \emph{don't} use DSFormer's semantic token and \emph{don't} include a frequency bias.

\section{PredCls Results}

\cref{tab:supp:predcls} contains \emph{mR@50} scores obtained with the Predicate Classification Protocol (\emph{PredCls}).
While the base (B) and large (L) variants perform very similar on \emph{PredCls}, their performance on \emph{SGDet} differs considerably.
Since the L variant contains a more capable segmentation model, it misses less relations in the \emph{SGDet} setting.
For \emph{PredCls}, the ground truth segmentation masks are provided and the difference in the quality of the segmentation model is irrelevant.

\begin{table}[h]
  \centering
  \caption{
    Scene graph performance, evaluated as \emph{mR@50} on the PSG dataset \scite{psg} using Predicate Classification (\emph{PredCls}) and Scene Graph Detection (\emph{SGDet}, aka \emph{SGGen}).
  }
  \label{tab:supp:predcls}
  \begin{tabular}{@{}lrr@{}}
    \toprule
    Method & PredCls & SGDet \\
    \midrule
    DSFlash-S & 39.27 & 26.62 \\
    DSFlash-B & 41.30 & 28.80 \\
    DSFlash-L & 41.69 & 30.90 \\
    \bottomrule
  \end{tabular}
\end{table}

\TODO{Perform some evaluation on Visual Genome.}

\section{Additional Backbone Ablation}
\label{sec:supp:backbone}

\begin{figure}[h]
  \centering
  \begin{subfigure}{0.49\linewidth}
    \includegraphics[width=\linewidth]{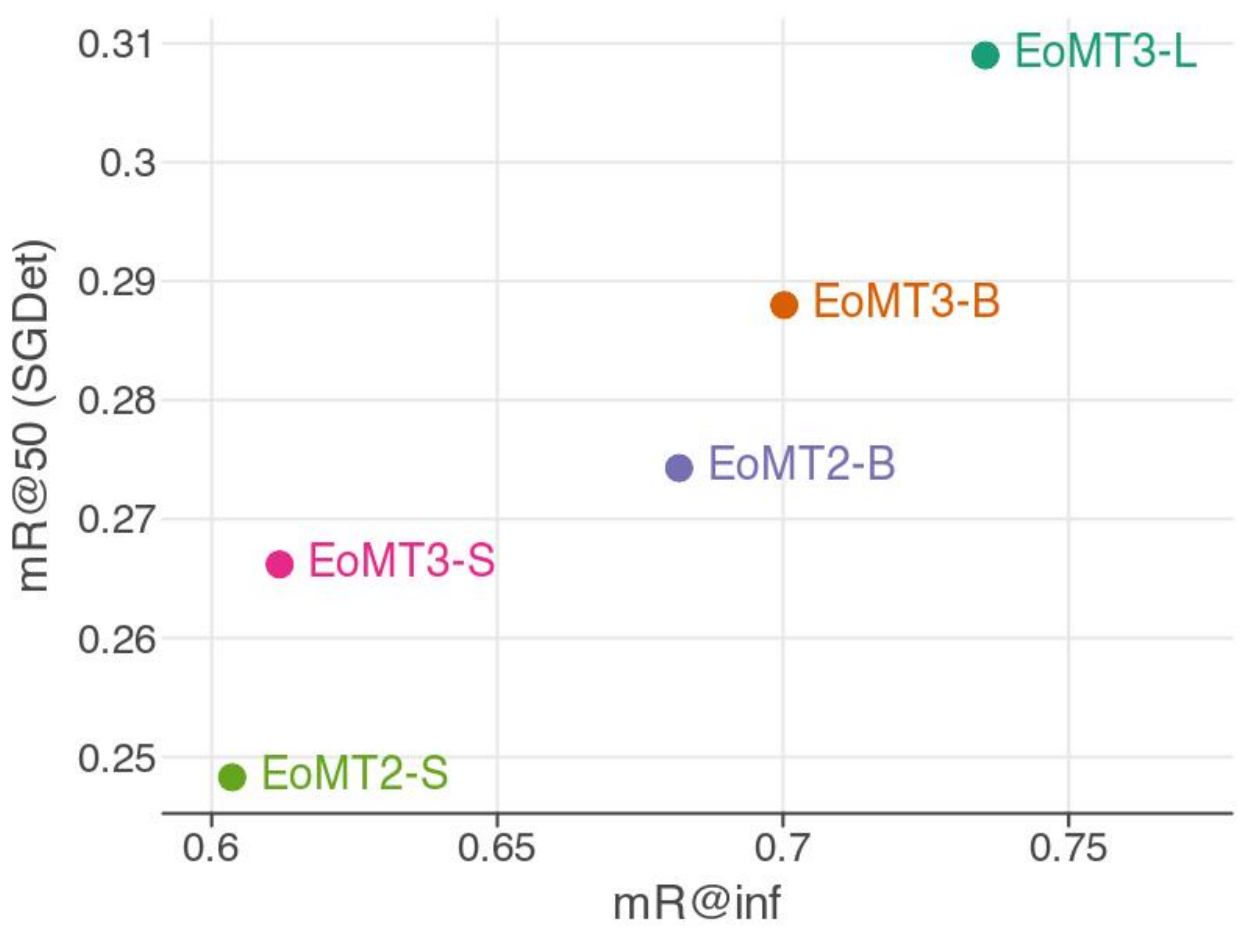}
  \end{subfigure}
  \hfill
  \begin{subfigure}{0.49\linewidth}
    \includegraphics[width=\linewidth]{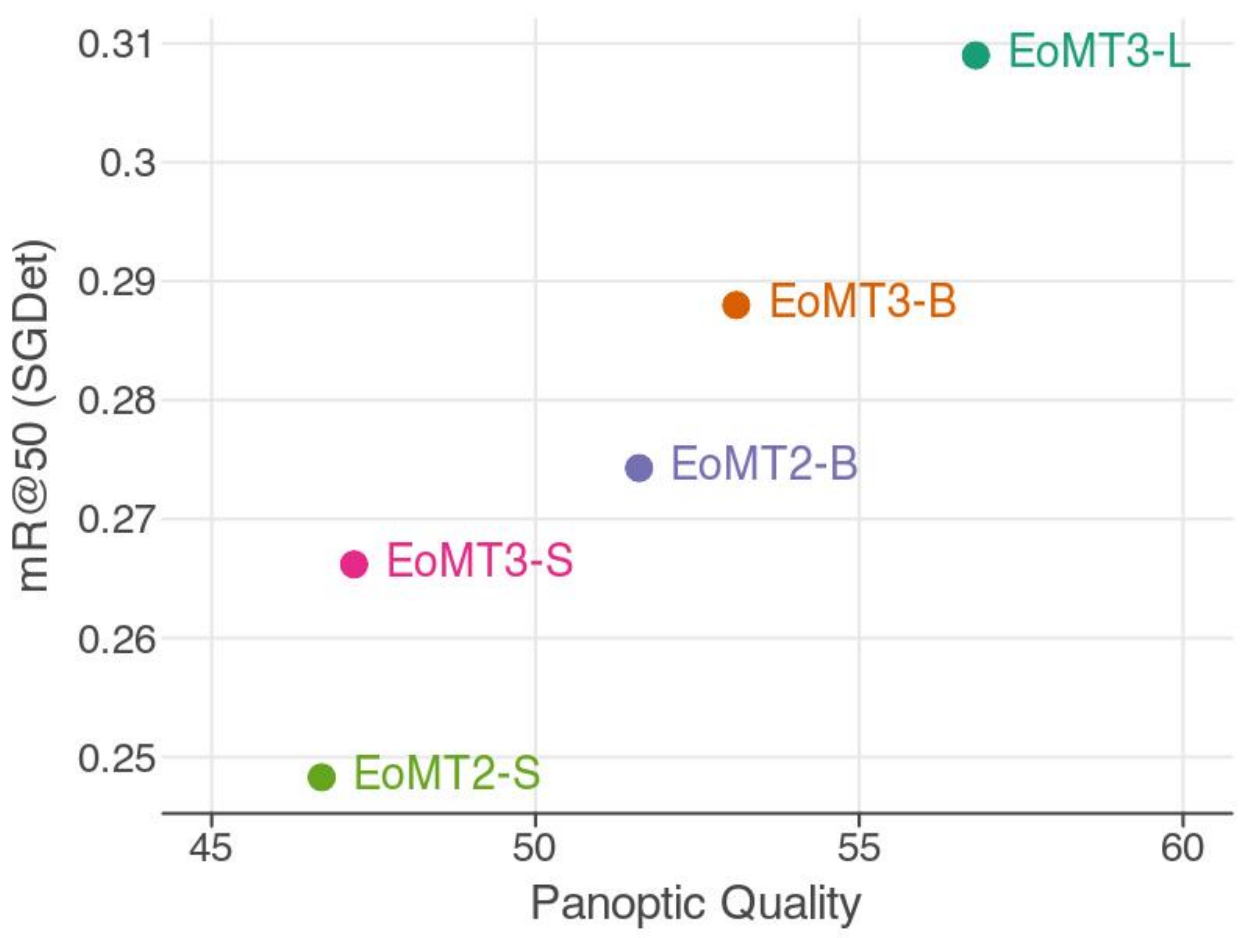}
  \end{subfigure}
  \caption{Correlation of scene graph performance (\emph{mR@50}) and backbone capabilities (measured in \emph{mR@inf} and panoptic quality).}
  \label{fig:supp:backbone}
\end{figure}

In addition to \cref{sec:bb-abl}, we provide additional backbone ablation results. \cref{fig:supp:backbone} shows how DSFlash's scene graph performance correlates with the panoptic quality (\emph{PQ}) of the segmentation model.
Using EoMT2 models (\ie pretrained with DINOv2) as the backbone for DSFlash results in slightly worse results than EoMT3 models.
We also observe a very strong correlation of 0.99 between \emph{mR@inf} and \emph{PQ}.

\begin{figure}[h]
  \centering
  \includegraphics[width=0.8\linewidth]{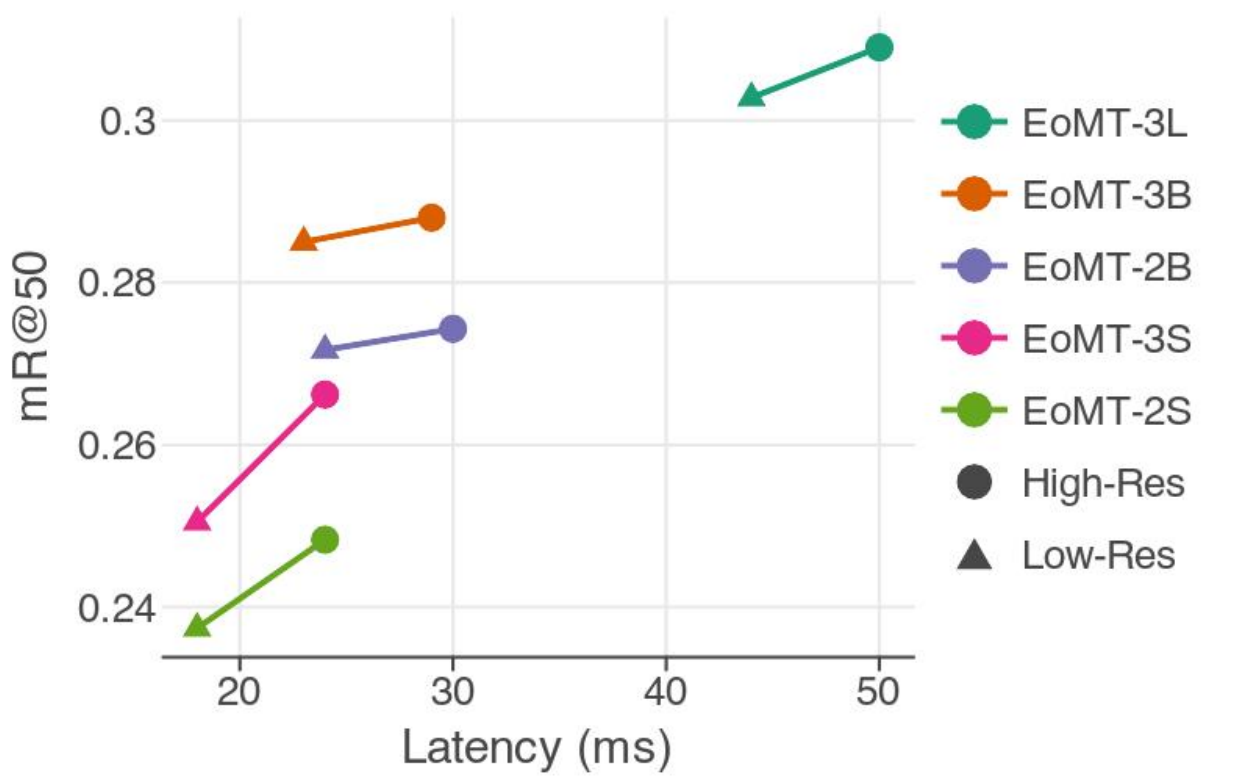}
  \caption{
    Comparison of the effect of low-resolution segmentation masks on the final performance and latency when using different segmentation backbones.
  }
\end{figure}

\section{Train with pruned patches/token merging}

\cref{tab:supp:tome-rps} shows RPS measurements on different hardware.
Considering the drop in \emph{mR@50}, there are only minor throughput gains on an H100 and RTX 3090.
On a GTX 1080, RPS improves by about 25\% compared to the baseline experiment when combining patch pruning and token merging.

\begin{table}[h]
  \centering
  \caption{Impact of Token Merging (\emph{ToMe}) and mask-based dynamic patch pruning (\emph{Prune}, \cref{sec:dyn-pruning}) on \model{}'s RPS, measured on a H100, RTX 3090, and GTX 1080 GPU.}
  \label{tab:supp:tome-rps}
  \begin{tabular}{@{}crrrrr@{}}
  \toprule
  Prune & ToMe & H100 & 3090 & 1080 & mR@50 \\
  \midrule
  \textcolor{red}{$\times$} & 0\% & 28,663 & 11,491 & 671 & 28.80 \\
  \textcolor{Green}{\checkmark} & 0\% & 26,734 & 11,933 & 727 & 26.67 \\
  \textcolor{Green}{\checkmark} & 30\% & 25,817 & 11,667 & 852 & 26.51 \\
  \textcolor{red}{$\times$} & 40\% & 26,298 & 11,154 & 823 & 25.82 \\
  \textcolor{red}{$\times$} & 50\% & 26,475 & 11,266 & 845 & 24.87 \\
  \textcolor{red}{$\times$} & 60\% & 27,062 & 11,559 & 810 & 21.93 \\
  \bottomrule
  \end{tabular}
\end{table}

\begin{figure}[h]
  \centering
  \includegraphics[width=\linewidth]{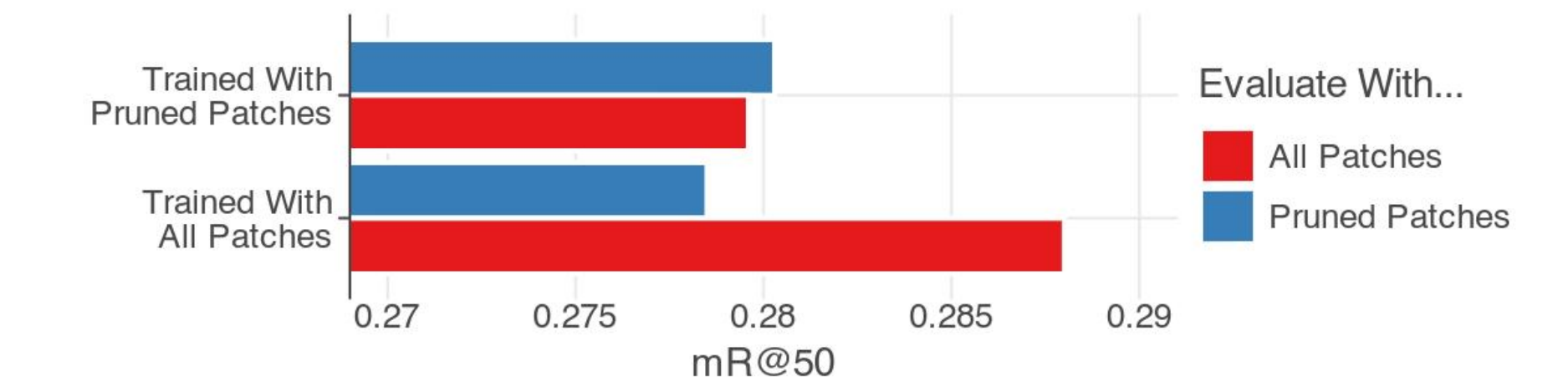}
  \caption{
    Comparison of a model trained with pruned patches and one without when pruning patches during evaluation.
  }
  \label{fig:supp:drop}
\end{figure}

Our results in \cref{sec:drop} investigate how patch pruning negatively impacts \emph{mR@50} performance of DSFlash.
However, when patch pruning is also enabled during training, DSFlash retains its scene graph generation capabilities even with pruned patches as shown in \cref{fig:supp:drop}, although at a lower overall performance.

\section{Qualitative Results}

\begin{figure}[h]
  \centering
  \includegraphics[width=0.8\linewidth]{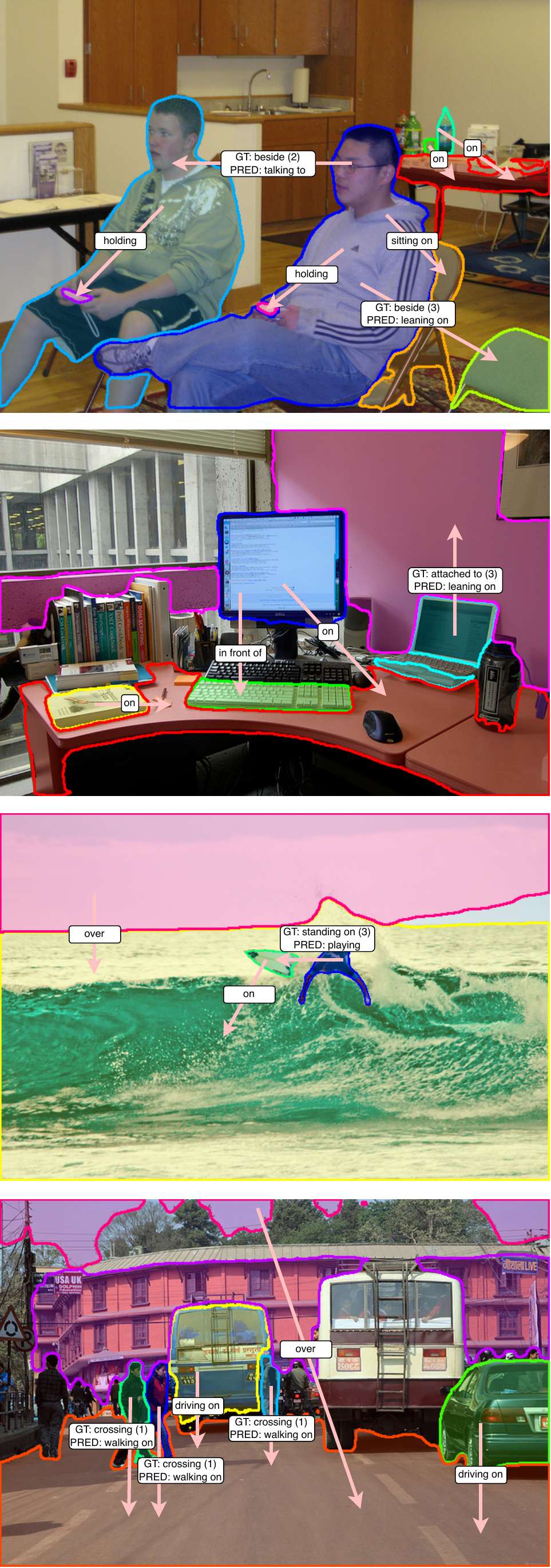}
  \caption{Qualitative results using DSFlash.}
  \label{fig:supp:qual}
\end{figure}

\cref{fig:supp:qual} shows some qualitative results.
If a label on an arrow has only one line, DSFlash correctly predicted the ground truth predicate.
If there are two lines, the bottom line shows DSFlash's prediction while the top line shows the ground truth predicate.
The number in parantheses shows at what (0-based) rank DSFlash positions the ground truth predicate.
For example, \texttt{GT: beside (2)} means that DSFlash's third guess would be \enquote{crossing}.

\begin{figure}[h]
  \centering
  \includegraphics[width=0.8\linewidth]{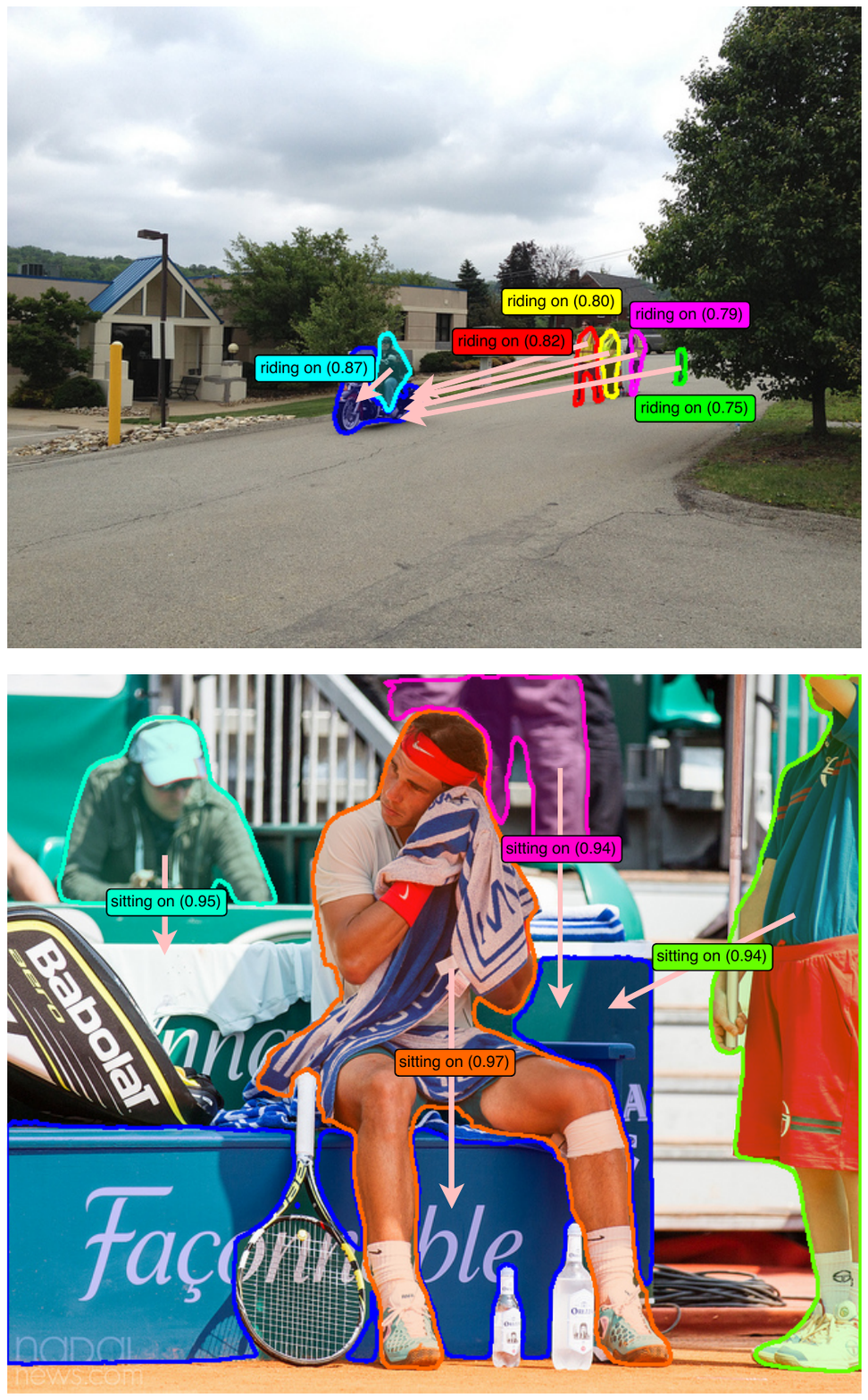}
  \caption{
    Failure cases when using DSFlash.
    Shown are predictions that are in the top 50 predictions for the respective image.
    The color in the boxes indicates to which subject the prediction is related to.
  }
  \label{fig:supp:fail}
\end{figure}

\cref{fig:supp:fail} shows some failure cases when using DSFlash.
In some cases, DSFlash confuses subjects and objects when predicting the relation.
We encourage future work to address this issue and improve DSFlash for these scenarios, for example using contrastive losses \scite{contrastive}.

{\small
\bibliographystyle{ieeenat_fullname}
\bibliography{main}
}

\end{document}